\documentclass[10pt,twocolumn,letterpaper]{article}

\usepackage{cvpr}              
\definecolor{cvprblue}{rgb}{0.21,0.49,0.74}
\usepackage[pagebackref,breaklinks,colorlinks,allcolors=cvprblue]{hyperref}

\usepackage[table]{xcolor}
\usepackage{multirow}
\usepackage{utfsym}
\usepackage{graphicx}
\usepackage{subcaption}

\def\paperID{} 
\def\confName{CVPR}
\def\confYear{2026}

\title{Weakly Supervised Video Anomaly Detection with Anomaly-Connected Components and Intention Reasoning}

\author{Yu Wang\thanks{Corresponding author: Yu Wang (email:yuwangtj@yeah.net)}  \quad Shengjie Zhao\\
School of Computer Science and Technology, Tongji University, Shanghai, China\\
}

\begin{document}
\maketitle
\begin{abstract}
Weakly supervised video anomaly detection (WS-VAD) involves identifying the temporal intervals that contain anomalous events in untrimmed videos, where only video-level annotations are provided as supervisory signals. However, a key limitation persists in WS-VAD, as dense frame-level annotations are absent, which often leaves existing methods struggling to learn anomaly semantics effectively. To address this issue, we propose a novel framework named LAS-VAD, short for \textbf{L}earning \textbf{A}nomaly \textbf{S}emantics for WS-\textbf{VAD}, which integrates anomaly-connected component mechanism and intention awareness mechanism. The former is designed to assign video frames into distinct semantic groups within a video, and frame segments within the same group are deemed to share identical semantic information. The latter leverages an intention-aware strategy to distinguish between similar normal and abnormal behaviors (e.g., taking items and stealing). To further model the semantic information of anomalies, as anomaly occurrence is accompanied by distinct characteristic attributes (i.e., explosions are characterized by flames and thick smoke), we additionally incorporate anomaly attribute information to guide accurate detection. Extensive experiments on two benchmark datasets, XD-Violence and UCF-Crime, demonstrate that our LAS-VAD outperforms current state-of-the-art methods with remarkable gains.
\end{abstract}    
\section{Introduction}
\label{sec:intro}

In recent years, video anomaly detection (VAD) has attracted growing attention owing to its broad practical applications, including intelligent surveillance \cite{fusedvision,ascnet}, multimedia content understanding \cite{sqlnet,lecvad}, and smart manufacturing \cite{aiad}. VAD seeks to identify temporal intervals containing anomalous events in arbitrarily long, untrimmed videos. A key challenge for VAD, however, lies in its dependence on expensive, dense annotations that precisely delineate the start and end of each anomalous event. To alleviate this reliance, weakly supervised VAD (WS-VAD) has been proposed. This paradigm enables training using only video-level annotations, which in turn streamlines the annotation workflow and improves the practical feasibility of VAD in real-world scenarios.

Most current WS-VAD approaches \cite{DDF,itc,lecvad,ovvad,vadclip,dmu,laa,ffpn,ssvad,mist,tsn} follow a standardized pipeline. First, the process involves extracting frame-level features using pre-trained models \cite{CLIP}. Next, these extracted features are fed into classifiers while incorporating a multiple instance learning (MIL) strategy \cite{mil} to identify anomaly. Despite significant advancements, there are still two limitations. On the one hand, the lack of frame-level supervisory signals results in difficulty in capturing anomaly semantic information. On the other hand, the distinction between abnormal and normal behaviors is ambiguous. For instance, actions like taking items and stealing exhibit little difference, with the only discrepancy lying in the speed of grasping.

To address the above limitations, we propose a novel framework named LAS-VAD, short for \textbf{L}earning \textbf{A}nomaly \textbf{S}emantics for WS-\textbf{VAD}. Specifically, to alleviate the problem of difficulty in capturing semantic information caused by the absence of frame-level supervisory signals, we develop a novel anomaly-connected component (ACC) approach. By calculating pairwise similarity between frames, ACC can divide frames into non-overlapping groups through the identification of connected components. Frame segments within the same group are deemed to share identical semantic information, which is utilized to guide frame-level learning. Besides, as anomaly occurrence is accompanied by distinct characteristic attributes, \textit{i.e.}, explosions are characterized by flames and thick smoke, we further incorporate anomaly attribute information to guide accurate detection.

To enhance the discriminability between normal and abnormal behaviors, we devise a novel intention-aware strategy. Specifically, normal and abnormal behaviors usually share similar appearance features, for example, taking items and stealing, but their intentions differ. To capture intentions, we first extract position features, velocity features and acceleration features, which are intended to better reason intentions. We then develop a novel intention prototype to store the semantic information of different intentions, along with a cross-intention contrastive learning mechanism to distinguish between distinct intentions.

The main contributions of this work are five-fold: 1) We propose LAS-VAD, a novel framework for weakly supervised video anomaly detection with anomaly-connected components and intention reasoning mechanisms. 2) To alleviate insufficient semantic supervisory signals caused by missing frame-level labels, we propose a novel anomaly-connected component approach to cluster video frames into distinct semantic groups, where frames in the same group share identical semantics. 3) To enhance the discriminability between normal and abnormal behaviors, we propose a novel intention awareness mechanism, utilizing intention prototypes and cross-intention contrastive learning, to distinguish between distinct intentions. 4) Since anomaly occurrence is accompanied by distinct characteristic attributes, we propose to incorporate anomaly attribute information to guide accurate detection. 5) Extensive evaluations on XD-Violence and UCF-Crime datasets have shown that our LAS-VAD achieves state-of-the-art performance.

\section{Related Work}
\label{sec:relatedWork}

\subsection{Vision-Language Pre-training}

Cross-modal vision-language comprehension \cite{vlm4d,umvl,bvll,sdo} is a core task requiring accurate alignment of representations between the visual and linguistic modalities. Mainstream approaches fall into two categories, namely joint and dual encoder architectures. Joint encoder-based methods \cite{abf,vinvl,vilbert} leverage a multi-modal encoder to enable fine-grained interactions between vision and language. Nevertheless, despite their powerful performance, these methods have a key limitation, \textit{i.e.}, they require processing each text-image pair separately during inference, resulting in substantial inefficiencies. In contrast, dual encoder-based approaches \cite{lecvad,ascnet,CLIP} adopt two separate encoders to extract visual and linguistic features. Recent years have seen significant progress driven by large-scale contrastive pre-training under this paradigm—progress that has greatly improved the performance of various multi-modal tasks, such as text-image retrieval \cite{cmirr,msrm}, visual question answering \cite{drivelm,fglim}, and video grounding \cite{univtg,crtua}. In this work, we adopt CLIP \cite{CLIP} and transfer it to the task of weakly supervised video anomaly detection.

\subsection{Weakly Supervised Video Anomaly Detection}
Weakly supervised video anomaly detection (WS-VAD) has garnered significant attention in recent years, due to its broad applicability \cite{sqlnet,ascnet,lecvad,mmh,aiad} and affordable manpower costs. Sultani \textit{et al.} \cite{ucf}, recognized as pioneers in this field, have compiled a large-scale video anomaly detection dataset with video-level annotations, and they have utilized a multiple instance ranking strategy for the localization of anomalous events. Subsequent efforts have focused on multiple aspects of improving WS-VAD performance. One research direction \cite{rtfm,gclnc,stms,icfs,sgtd,dmu} has explored capturing temporal relationships among video segments, achieved via graph structures \cite{gclnc}, self-attention strategies \cite{dmu,rtfm}, and transformers \cite{sgtd,umil,stms}. These methods offer a deeper understanding of how distinct video parts interact, a factor crucial for accurate anomaly detection. The other research direction has been to explore self-training schemes \cite{MEL,tpng,ecup,vadstp,mist} that generate snippet-level pseudo-labels, which are then utilized to iteratively refine anomaly scores. This process aids in enhancing anomaly detection accuracy by offering more granular feedback. Additionally, Sapkota \textit{et al.}\cite{bnsv} and Zaheer \textit{et al.} \cite{cawst} have sought solutions to mitigate the impact of false positives.

Recently, research efforts to utilize multi-modal knowledge for enhancing WS-VAD performance have surged. Notable examples include PEL4VAD \cite{pel4vad}, VadCLIP \cite{vadclip}, and ITC \cite{itc}, which inject text clues corresponding to anomaly-event categories into WS-VAD, whereas PE-MIL \cite{pe-mil} and MACILSD \cite{MACIL} introduce audio features to strengthen the capability of WS-VAD. Such innovations have delivered promising outcomes. Wu \textit{et al.} \cite{tvar} have taken a step further to develop a framework enabling the retrieval of video anomalies using language queries or synchronous audio. Wang \textit{et al.} \cite{lecvad} first propose to learn event completeness with Gaussian priors for WS-VAD. $\pi$-VAD \cite{piVAD} leverages five additional modalities to augment RGB features of images.

Besides, pre-trained vision-language models have also emerged as a powerful tool to extract robust representations \cite{tgad,vadclip,tpng,reflip}. For instance, Wu \textit{et al.} propose an open-vocabulary video anomaly detection paradigm \cite{ovvad}, designed to identify both known and novel anomalies in open-world scenarios. Huang \textit{et al.} \cite{MEL} further devise a multi-scale evidential vision-language model to achieve open-world video anomaly detection. Yang \textit{et al.} \cite{tpng} employ pre-trained CLIP for generating reliable pseudo-labels. Lv \textit{et al.} \cite{umil} have designed an unbiased multiple-instance strategy for learning invariant representations. Wang \textit{el al.} \cite{federatedVAD} design a prompt generator driven by both global and local contexts, which achieves global generalization and local personalization. STPrompt \cite{stPrompt} acquires spatio-temporal prompt embeddings by leveraging pre-trained vision-language models.

Despite significant progress, previous methods struggle to learn anomaly semantics due to the absence of frame-level annotations. In contrast, we propose LAS-VAD, a novel framework for WS-VAD with anomaly-connected components and intention reasoning mechanisms.
\begin{figure*}
\centering 
\includegraphics[width=0.78\textwidth]{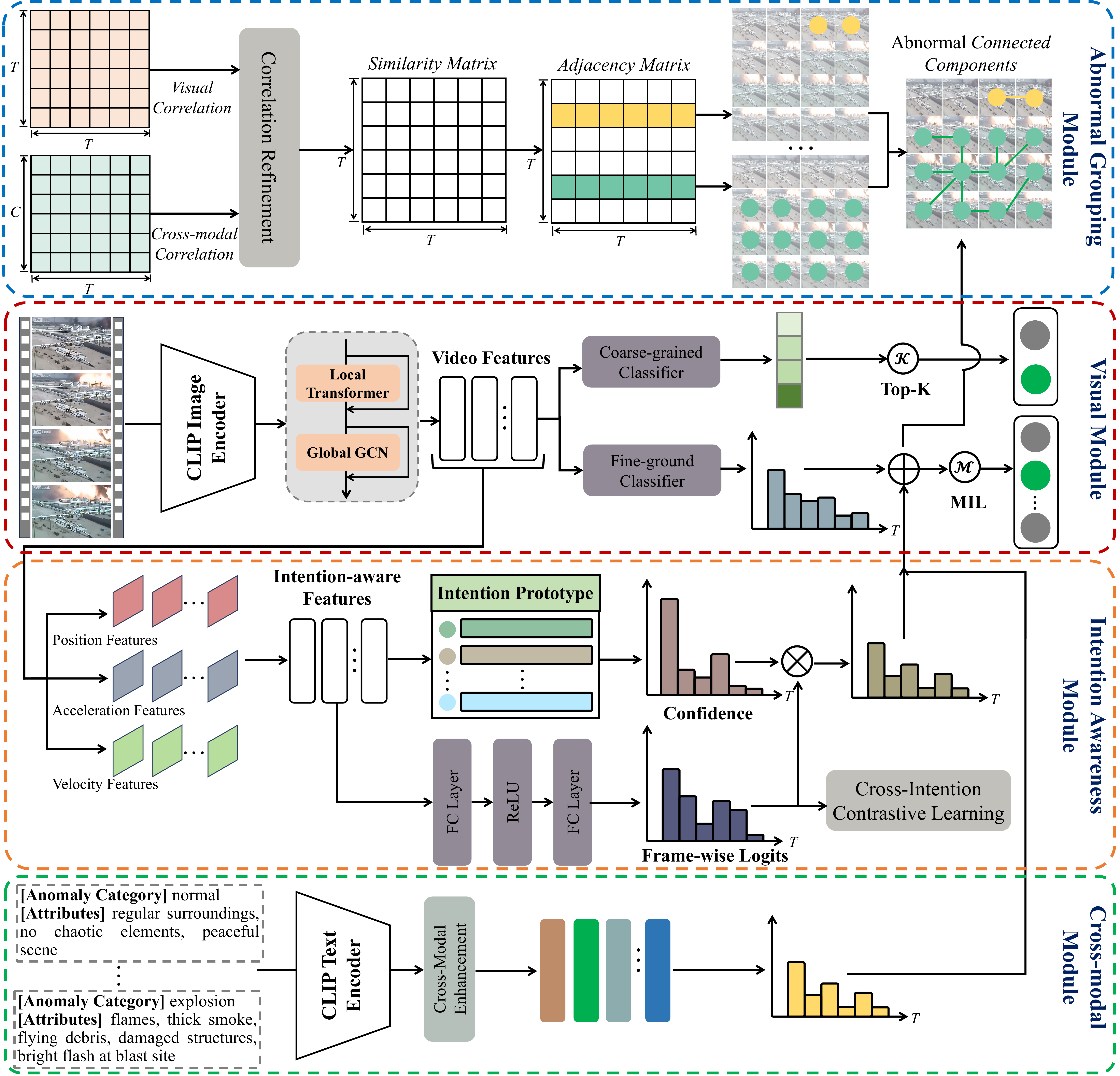} 
\vspace{-0.5em}
\caption{Overview of the proposed LAS-VAD. We integrate an intention awareness mechanism and anomaly-connected component mechanism into WS-VAD to learn object intentions and more discriminative semantics. Additionally, attribute information of anomaly categories is introduced to facilitate accurate anomaly detection.}
\vspace{-1em}
\label{fig:framework} 
\end{figure*}

\section{Methodology}
\label{sec:method}

\subsection{Overview}
\label{sec:overview}

\textbf{Problem Definition.} Consider a dataset of $n$ videos $\mathcal{V}=\{v_i\}_{i=1}^n$ with different annotation granularities. Each video $v_i$ has both coarse-grained and fine-grained video-level labels, with $y_i \in \{0,1\}$ and $g_i \in \{0, ..., C\}$ corresponding to these two types respectively, and $C$ here denotes the number of anomaly categories. If $v_i$ is considered normal, then $y_i = 0$ and $g_i = 0$, otherwise $y_i = 1$ and $g_i$ is assigned a specific category. For coarse-grained WS-VAD, the task is to estimate the likelihood of an anomaly being present in each snippet. Fine-grained WS-VAD, however, calls for predicting specific anomalous event instances that are denoted as $\{s_j, e_j, a_j, u_j\}$. Here, $s_j$ and $e_j$ mark the start and end timestamps of the $j$-th anomaly instance, $a_j$ stands for its corresponding anomaly category, and $u_j$ indicates the prediction confidence.

\noindent\textbf{Framework.}
This paper proposes an innovative framework, LAS-VAD, for WS-VAD, as illustrated in the Fig. \ref{fig:framework}. The core idea is to strengthen the semantic awareness capability. To this end, we develop a novel anomaly-connected component approach for partitioning video frames into distinct semantic groups, where frames within the same group hold consistent semantics. This allows us to guide the frame-level learning using the semantics of the corresponding group. To better capture anomaly semantics, we note that anomalies are accompanied by distinct characteristic attributes, \textit{e.g.}, flames and thick smoke define explosions, and thus integrate anomaly attribute features to enhance detection accuracy. In addition, to boost the discriminability between normal and abnormal behaviors, we devise a novel intention awareness mechanism that employs intention prototypes and cross-intention contrastive learning to distinguish distinct intentions. In detail, we adopt a pre-trained image encoder of CLIP to extract visual features $X_{video} \in \mathcal{R}^{T \times D}$, in which $T$ is the total number of frames and $D$ denotes the dimension of features. To encode temporal dependencies, frame-level features are partitioned into overlapping windows of uniform length. A local transformer layer with constrained attention ranges \cite{swin} is subsequently applied, thereby precluding information exchange across these windows. The enhanced feature $X_h \in \mathcal{R}^{T \times D}$ is generated through this procedure. Next, to further model global temporal dependencies, we deploy a GCN module on $X_h$. Specifically, in line with the configuration outlined in \cite{vadclip}, the global temporal correlations among features are modeled according to their similarity. This process can be summarized as follows:
\begin{equation}
    \begin{split}
        A &= \frac{X_hX_h^T}{||X_h||_2 \cdot ||X_h||_2},\\
        X_{f} &= GELU(Softmax(A)X_lW),
    \end{split}
\end{equation}
where $A$ refers to the adjacency matrix, $W$ denotes the learnable weight, and $X_{f}$ corresponds to the enhanced video feature. The Softmax operation is employed to guarantee that the sum of elements in each row equals unity. Subsequently, a pre-trained text encoder is employed to extract linguistic features for all anomaly event categories, denoted as $X_{lang} \in \mathcal{R}^{(C+1) \times D}$. To better capture the semantic information of anomalies, we note that anomalies exhibit distinct attributes (\textit{e.g.}, explosions are marked by flames and thick smoke) and further integrate anomaly attribute information to enhance detection accuracy. To further model the semantic information of anomalies, as anomaly occurrence is accompanied by distinct characteristic attributes (\textit{i.e.}, explosions are characterized by flames and thick smoke), we additionally incorporate anomaly attribute information to guide accurate detection. Specifically, we leverage an off-the-shelf LLM to generate textual descriptions of each anomaly category $E_i$, where $ i \in \{0,...,C\}$. Subsequently, we employ the pre-trained text encoder to extract features from these descriptions, which are denoted as $X_{aux} \in \mathcal{R}^{(C+1) \times D}$. We concatenate $X_{lang}$ and $X_{aux}$ to acquire the final textual features $X_{text}$ for each anomaly category. We further compute the cross-modal cosine similarity between $X_{text}$ and $X_{f}$ to obtain frame-level fine-grained classification scores $q^l \in \mathcal{R}^{T \times (C+1)}$. 

Meanwhile, we apply a fully-connected layer to $X_{f}$, followed sequentially by a Sigmoid operation for binary classification in category-agnostic anomaly detection, and a Softmax operation for multi-class classification in category-aware anomaly detection, generating frame-level predictions $\textit{q}^{b} \in \mathcal{R}^{T \times 2}$ and $\textit{q}^{m} \in \mathcal{R}^{T \times (C+1)}$, respectively. 

For the $c$-th category in $\textit{q}^{b}$, we calculate the average of the top-$K$ scores along the temporal dimension, as follows:
\begin{equation}
    \begin{split}
        p^b_c = \max_{U \subseteq \{1,...,T\}\atop |U|=K}\frac{1}{K}\sum_{j \in U}\textit{q}^{b}_{jc},
    \end{split}
\end{equation}
where $U$ is the index set of top-$K$ frames, and $p^b_c$ indicates the category-agnostic prediction. Then, a binary cross-entropy loss is applied between $p^b$ and the ground truth $y$:
\begin{equation}
    \begin{split}
        \mathcal{L}_{ags} = -\sum_{c \in \{0, 1\}} y_c \log(p^b_c).
    \end{split}
\end{equation}

Although $\textit{q}^{m}$ encodes category-aware anomaly clues, it suffers from insufficient discriminative ability and limited intention awareness due to the lack of frame-wise annotations. To this end, we propose an intention awareness module (IAM) as in Section \ref{sec:iam} and a novel anomaly-connected component module (ACC) as in Section \ref{sec:acc}. In general, to enhance the discriminability between normal and abnormal behaviors, IAM attempts to establish intention prototypes and adopt a cross-intention contrastive learning mechanism to distinguish between distinct intentions, yielding a contrastive loss $\mathcal{L}_{cst}$ and frame-level predictions $q^a \in \mathcal{R}^{T \times (C+1)}$. Finally, we define frame-level fine-grained classification logits as $p^f = \frac{1}{3}(q^m + q^a + q^l)$. $p^f$ is then integrated into video-level scores $p^v$ via a MIL mechanism. Specifically, for the $c$-th category, the $p^v_c$ is computed as follows:
\begin{equation}
    \begin{split}
        p^v_c = \max_{U \subseteq \{1,...,T\} \atop |U|=K}\frac{1}{K}\sum_{t \in U}\textit{p}^{f}_{t}.
    \end{split}
\end{equation}
Next, a cross-entropy constraint is employed to supervise fine-grained anomaly classification, as follows:
\begin{equation}
    \begin{split}
        \mathcal{L}_{fg} = -\sum_{c=0}^C g_c \log(p_c^v).
    \end{split}
\end{equation}

Meanwhile, ACC module partitions video frames into distinct semantic groups, where frames within the same group are deemed to share identical semantics, yielding frame-level pseudo labels $g \in \mathcal{R}^{T \times (C+1)}$. These pseudo labels are further used to aid the learning of $\textit{q}^{m}$ by:
\begin{equation}
    \begin{split}
        \mathcal{L}_{aux} = \frac{1}{T}\sum_{t=1}^T \sum_{c=0}^C |{g}[t,c] - q^m[t,c]|.
    \end{split}
\end{equation}

Therefore, the base optimization loss of LAS-VAD is:
\begin{equation}
    \begin{split}
       \mathcal{L}_{base} = \mathcal{L}_{ags} + \mathcal{L}_{fg} + \mathcal{L}_{aux}.
    \end{split}
\end{equation}

Besides, to ensure the consistency between category-agnostic and category-aware predictions, \textit{i.e.}, $p^b[t,1]$ and $1 - p^f[t,0]$, $\ell_1$ regularization is introduced, formulated as:
\begin{equation}
    \begin{split}
       \mathcal{L}_{reg} = \frac{1}{T}\sum_{t=1}^T |1 - p^f[t,0] - p^b[t,1]|.
    \end{split}
\end{equation}

Overall, the optimization loss of LAS-VAD is:
\begin{equation}
    \begin{split}
       \mathcal{L}_{all} = \mathcal{L}_{base} + \lambda\mathcal{L}_{reg}.
    \end{split}
\end{equation}
where $\lambda$ is used to balance different terms.

\subsection{Intention Awareness Mechanism}
\label{sec:iam}

To enhance the discriminability between normal and abnormal behaviors, we propose a novel intention-aware strategy in this section. Specifically, normal and abnormal behaviors frequently share similar appearance features. For instance, the behaviors of ``taking objects" and ``stealing" exhibit similar appearance but distinct intentions, a difference reflected in their object-taking speed, \textit{i.e.}, the speed in theft cases is significantly faster than that in normal situations. For this purpose, we seek to reason about the behavioral intention to better capture anomalies.

Based on the video feature $X_f$, we first generate position features $X_p \in \mathcal{R}^{T \times (D/3)} $, velocity features $X_v \in \mathcal{R}^{T \times (D/3)}$ using the difference between adjacent positions, and acceleration features $X_a \in \mathcal{R}^{T \times (D/3)}$ using the difference between adjacent velocities. The implementation of this procedure is detailed below:
\begin{equation}
    \begin{split}
        X_p = fc(X_f), \quad X_v^{diff} = |X_p[t,:] - X_p[t-1,:]|,\\
        X_v = Sigmoid(Conv(X_v^{diff}))\times X_v^{diff},\\
        \quad X_a^{diff} = |X_v[t,:] - X_v[t-1,:]|,\\
        X_a = Sigmoid(Conv(X_a^{diff}))\times X_a^{diff},\\
    \end{split}
\end{equation}
where $fc$ denotes the fully-connected layer, $Conv$ is a convolution operation with a kernel size of 3, and zero elements are padded at the start and end positions of $X_v$ and $X_a$ to ensure that their temporal dimension remains $T$. Then, we concatenate $X_p$, $X_v$, and $X_a$ along the feature dimension, forming intention-aware features $X_{int} \in \mathcal{R}^{T \times D}$.

Utilizing intention-aware features $X_{int}$, we acquire frame-wise logits $q_{int} \in \mathcal{R}^{T \times C}$ through two fully-connected layers with a ReLU activation. Meanwhile, we establish intention prototypes $Z \in \mathcal{R}^{(C+1) \times D}$ to retain semantic prototypes of different categories. On the one hand, for the label corresponding to the dimension where the maximum value lies at each time-step in $q_{int}$, we calculate the cosine similarity between the prototype of this label from $Z$ and its feature from $X_{int}$, yielding the confidence $w_{int} \in \mathcal{R}^{T \times 1}$ for the predicted logits $q_{int}$. We then obtain the weighted frame-wise scores $q_{a} = Softmax(q_{int} \times w_{int})$. On the other hand, we also update the intention prototypes in a momentum manner. In detail, for the anomaly category $c$, the corresponding features with classification scores in $q_{a}$ greater than a threshold $\alpha$ are selected, and their mean value $X_{center}^c$ is used to update the corresponding class prototype as $Z[c,:] = (1-\beta) \times Z[c,:] + \beta \times X_{center}^c$, where $\beta$ denotes the momentum coefficient.

Furthermore, we propose a cross-intention contrastive learning strategy aimed at explicitly mitigating the issue of intention confusion. The core principle resides in mining hard-to-distinguish frame pairs with different intentions and easily confusable frame pairs with the same intentions. Specifically, for each frame feature $X_{int}^t$, we select frames as positive samples $S_{pos}^t$ with the lowest similarity among samples of the same category, and as negative samples $S_{neg}^t$ with the top-$M$ highest similarity among different intentions. Subsequently, the infoNCE loss is adopted to constrain the intention distribution:
\begin{equation}
    \begin{split}
        \mathcal{L}_{cst} = -\frac{1}{T}\sum_{t=1}^T\log\frac{\exp(X_{int}^t \cdot S_{pos}^t)}{\sum_{i=1}^M \exp(X_{int}^t \cdot S_{neg}^t)}.
    \end{split}
\end{equation}

\subsection{Anomaly-Connected Component Mechanism}
\label{sec:acc}
To alleviate insufficient semantic supervisory signals arising from the lack of frame-level annotations, we propose a novel anomaly-connected component approach to partition video frames into distinct semantic groups, with frames within the same group sharing consistent semantics. Specifically, given video features $X_f \in \mathcal{R}^{T \times D}$, we compute the pairwise similarity between each frame:
\begin{equation}
    \begin{split}
        \mathcal{A}_v = (\frac{X_f \cdot X_f^T}{||X_f||_{dim=1} \cdot ||X_f||_{dim=1}}) \in \mathcal{R}^{T \times T}.
    \end{split}
\end{equation}
However, $\mathcal{A}_v$ only captures the visual similarity of frames, which tends to be biased. To address this issue, we further leverage cross-modal similarity $q^l \in \mathcal{R}^{T \times (C+1)}$ to rectify such bias. The core principle is that for any $i$-th and $j$-th frames, if both exhibit high semantic similarity to the same text category,  their similarity is enhanced; otherwise, it is weakened. This process is formulated as:
\begin{equation}
    \begin{split}
        \mathcal{\hat{A}}_w[i,j] = \mathcal{A}_v[i,j] \cdot (1 + \eta \cdot \max_c\min(q^l[i,c],q^l[j,c])),
    \end{split}
\end{equation}
where $\eta$ denotes the degree of rectification, $\max_c\min(q^l[i,c],q^l[j,c])$ measures the textual semantic consistency between the two frames, and $\mathcal{\hat{A}}$ is the enhanced similarity map. We then convert it into a binary map $\mathcal{\hat{A}}$ based on a threshold $\tau$: $\mathcal{A} = (\mathcal{\hat{A}} > \tau) \in \mathcal{R}^{T \times T}.$
Here, each frame is regarded as a vertex, and $\mathcal{A}$ is treated as an adjacency matrix that characterizes inter-vertex connectivity. On this basis, we construct a graph, and the problem of identifying frames with consistent semantics is converted into the task of detecting all connected components in the graph via the adjacency matrix $\mathcal{A}$. To this end, we employ the Depth-First Search algorithm to obtain $r$ distinct connected components $B_1$, $B_2$,..., and $B_r$ from $\mathcal{A}$. 

For any $B_i$, which contains $u$ video frames, we get the feature prototype $X_{ac}^i$ by averaging their corresponding features within $X_f$, and semantic category $y_{proto}^i$ by averaging their logits within $p^f$. Subsequently, we compute the similarity between each frame's feature and all connected components' feature prototypes, and assign the semantic category of the most similar one as the frame's category label, yielding frame-level pseudo-labels $g \in \mathcal{R}^{T \times (C+1)}$.

\subsection{Inference}
\label{sec:inference}

During inference, coarse-grained WS-VAD calculates the average value of $1 - p^f[t,0]$ and $p^b[t,1]$ as the anomaly confidence for the $t$-th frame. For fine-grained WS-VAD, a two-step thresholding strategy is used to generate anomaly instances. First, anomaly categories with video-level activations exceeding the predefined threshold  $\theta_{v}$ are retained. Then, for each retained category, snippets with fine-grained matching scores in $p^f$ above $\theta_{s}$ are selected as candidates. These temporally consecutive candidates are merged into anomaly instances. Following AutoLoc \cite{autoloc}, each instance’s outer-inner score derived from $p^f$ serves as its confidence score, based on which non-maximal suppression (NMS) is applied to eliminate redundant proposals.

\section{Experiments}
\label{sec:sep}

\begin{table}[]
\renewcommand{\arraystretch}{1} 
\resizebox{0.48\textwidth}{!}{
\begin{tabular}{c|c|l|c|c}
\toprule
Supervision  & Modality   & \multicolumn{1}{c|}{Methods}  & Feature   & AP(\%) \\ 
\midrule
\multirow{1}{*}{Unsupervised}  & \multirow{1}{*}{RGB+Audio}  & LTR  \cite{ltr}    & I3D+VGGish & 30.77  \\ 
\midrule
\multirow{25}{*}{\begin{tabular}[c]{@{}c@{}}Weakly \\ Supervised\end{tabular}} & \multirow{6}{*}{RGB+Audio} & CTRFD \cite{CTRFD}  & I3D+VGGish & 75.90  \\
&   & WS-AVVD \cite{avvd}        & I3D+VGGish & 78.64  \\
&   & ECU  \cite{ecup}           & I3D+VGGish & 81.43  \\
&   & DMU \cite{dmu}             & I3D+VGGish & 81.77  \\
&   & MACIL-SD \cite{MACIL}      & I3D+VGGish & 83.40  \\ 
&   & PE-MIL \cite{pe-mil}       & I3D+VGGish & 88.21  \\ 
\cline{2-5} 
& \multirow{19}{*}{RGB}       & RAD  \cite{ucf}    & C3D        & 73.20  \\
&   & RTFML \cite{rtfm}    & I3D        & 77.81  \\
&   & ST-MSL \cite{stms}   & I3D        & 78.28  \\
&   & LA-Net \cite{lanet}   & I3D        & 80.72 \\
&   & CoMo \cite{como}     & I3D        & 81.30  \\
&   & DMU \cite{dmu}     & I3D        & 82.41  \\
&   & PEL4VAD \cite{pel4vad}     & I3D        & 85.59  \\
&   & PE-MIL \cite{pe-mil} & I3D & 88.05  \\ 
&   & $\pi$-VAD \cite{piVAD} & I3D & 85.37  \\ 
&   & LEC-VAD \cite{lecvad}  & I3D & 88.47      \\
&   & \cellcolor[HTML]{C0C0C0} LAS-VAD (Ours)  & \cellcolor[HTML]{C0C0C0} I3D & \cellcolor[HTML]{C0C0C0} \textbf{89.96}      \\
 \cline{3-5} 
&   & OVVAD  \cite{ovvad}  & CLIP       & 66.53  \\
&   & CLIP-TSA \cite{cliptsa} & CLIP    & 82.19  \\
&   & IFS-VAD \cite{icfs} & CLIP        & 83.14 \\
&   & TPWNG \cite{tpng}   & CLIP        & 83.68 \\
&   & VadCLIP \cite{vadclip} & CLIP     & 84.51  \\
&   & ITC  \cite{itc}    & CLIP         & 85.45  \\
&   & ReFLIP \cite{reflip}  & CLIP      & 85.81  \\
&   & LEC-VAD   \cite{lecvad}   &  CLIP       &  86.56   \\
&   & \cellcolor[HTML]{C0C0C0} LAS-VAD (Ours)  & \cellcolor[HTML]{C0C0C0} CLIP       & \cellcolor[HTML]{C0C0C0} \textbf{87.92}      \\
\bottomrule
\end{tabular}
}
\caption{Coarse-grained comparisons on XD-Violence.}
\label{tab:cg_xd}
\vspace{-1.7em}
\end{table}

\begin{table}[]
\renewcommand{\arraystretch}{1} 
\resizebox{0.48\textwidth}{!}{
\begin{tabular}{c|l|c|c}
\toprule
Supervision   & \multicolumn{1}{c|}{Methods}  & Feature   & AUC(\%) \\ 
\midrule
\multirow{3}{*}{Unsupervised}  & \multicolumn{1}{l|}{LTR \cite{ltr}} & I3D+VGGish & 50.60  \\
 &  \multicolumn{1}{l|}{GODS \cite{gods}}    & I3D   & 70.46  \\ 
 &  \multicolumn{1}{l|}{GCL  \cite{gcl}}    & I3D   & 71.04  \\ 
\midrule
\multirow{29}{*}{\begin{tabular}[c]{@{}c@{}}Weakly \\ Supervised\end{tabular}}  & \multicolumn{1}{l|}{TCN-CIBL \cite{CIBL}}    & C3D     & 78.66  \\
 & \multicolumn{1}{l|}{GCN-Anomaly \cite{gclnc}} & C3D     & 81.08  \\
 & \multicolumn{1}{l|}{GLAWS  \cite{claws}}     & C3D     & 83.03  \\
 & LEC-VAD \cite{lecvad}  & C3D    &  84.75  \\
  & \cellcolor[HTML]{C0C0C0} LAS-VAD (Ours) & \cellcolor[HTML]{C0C0C0} C3D    & \cellcolor[HTML]{C0C0C0} \textbf{86.04}  \\
 \cline{2-4} 
 & \multicolumn{1}{l|}{CTRFD  \cite{CTRFD}}    & I3D     & 84.89  \\ 
 & \multicolumn{1}{l|}{LA-Net  \cite{lanet}}   & I3D     & 85.12 \\
 & \multicolumn{1}{l|}{ST-MSL  \cite{stms}}    & I3D     & 85.30  \\
 & \multicolumn{1}{l|}{IFS-VAD \cite{icfs}}    & I3D     & 85.47 \\
 & \multicolumn{1}{l|}{NG-MIL \cite{ngmil}}    & I3D     & 85.63 \\
 & \multicolumn{1}{l|}{CLAV  \cite{laa}}       & I3D     & 86.10 \\
 & \multicolumn{1}{l|}{ECU   \cite{ecup}}      & I3D     & 86.22  \\
 & \multicolumn{1}{l|}{DMU   \cite{dmu}}       & I3D     & 86.75 \\
  & \multicolumn{1}{l|}{PE-MIL \cite{pe-mil}}  & I3D     & 86.83 \\
 & \multicolumn{1}{l|}{MGFN \cite{mgfn}}       & I3D     & 86.98 \\
 & \multicolumn{1}{l|}{$\pi$-VAD \cite{piVAD}}   & I3D     & 90.33 \\
&  LEC-VAD \cite{lecvad} & I3D    & 88.21  \\
& \cellcolor[HTML]{C0C0C0} LAS-VAD (Ours) & \cellcolor[HTML]{C0C0C0} I3D    & \cellcolor[HTML]{C0C0C0} \textbf{91.05}  \\
  \cline{2-4} 
 & \multicolumn{1}{l|}{Ju \textit{et al.} \cite{jupvl}} & CLIP & 84.72 \\
 & \multicolumn{1}{l|}{OVVAD  \cite{ovvad} }    & CLIP    & 86.40  \\
 & \multicolumn{1}{l|}{IFS-VAD  \cite{icfs} }  & CLIP    & 86.57 \\
 & \multicolumn{1}{l|}{UMIL \cite{umil}  }    & CLIP    & 86.75  \\
 & \multicolumn{1}{l|}{CLIP-TSA  \cite{cliptsa}}  & CLIP    & 87.58  \\
 &  \multicolumn{1}{l|}{TPWNG \cite{tpng}}   & CLIP   & 87.79 \\
 & \multicolumn{1}{l|}{VadCLIP  \cite{vadclip}}   & CLIP    & 88.02  \\
 & \multicolumn{1}{l|}{STPrompt \cite{stPrompt}}   & CLIP    & 88.08 \\
 & \multicolumn{1}{l|}{ReFLIP \cite{reflip} }    & CLIP    & 88.57 \\
 & \multicolumn{1}{l|}{ITC    \cite{itc} }    & CLIP    & 89.04  \\
 & LEC-VAD \cite{lecvad} & CLIP    &  89.97  \\
  & \cellcolor[HTML]{C0C0C0} LAS-VAD (Ours)    & \cellcolor[HTML]{C0C0C0} CLIP    & \cellcolor[HTML]{C0C0C0} \textbf{90.86}  \\

\bottomrule
\end{tabular}
}
\vspace{-0.5em}
\caption{Coarse-grained comparisons on UCF-Crime.}
\label{tab:cg_ucf}
\vspace{-1em}
\end{table}

\begin{table}[]
\renewcommand{\arraystretch}{1.05} 
\resizebox{0.48\textwidth}{!}{
\begin{tabular}{l|cccccc}
\toprule
\multirow{2}{*}{Methods}        & \multicolumn{6}{c}{mAP@IoU}   \\ 
\cline{2-7} 
& 0.1       & 0.2      & 0.3     & 0.4     & 0.5     & AVG    \\ 
\midrule
RAD   \cite{ucf}      & 22.72   & 15.57   & 9.98    & 6.20   & 3.78    & 11.65   \\
AVVD  \cite{avvd}   & 30.51   & 25.75   & 20.18   & 14.83  & 9.79    & 20.21   \\
VadCLIP \cite{vadclip} & 37.03   & 30.84   & 23.38   & 17.90  & 14.31   & 24.70   \\
ITC   \cite{itc}       & 40.83   & 32.80   & 25.42   & 19.65  & 15.47   & 26.83    \\
ReFLIP \cite{reflip} & 39.24   & 33.45   & 27.71   & 20.86  & 17.22   & 27.36 \\ 
LEC-VAD \cite{lecvad}   & 48.78  & 40.94   & 34.28   & 28.02   & 23.68  & 35.14 \\
\rowcolor[HTML]{C0C0C0} 
LAS-VAD (Ours)   & \textbf{50.18}  & \textbf{44.09}   & \textbf{36.00}   & \textbf{29.51}   & \textbf{24.65}  & \textbf{36.89} \\
\bottomrule
\end{tabular}
} 
\vspace{-0.5em}
\caption{Fine-grained comparisons on XD-Violence.}
\label{tab:fg_xd}
\vspace{-1.6em}
\end{table}

\subsection{Experimental Settings}
\textbf{Datasets.} \textbf{UCF-Crime} \cite{ucf} comprises 1900 untrimmed videos, which span 13 distinct categories of anomalous events. These videos are derived from surveillance recordings captured in a range of indoor environments and street-based scenarios. For experimental consistency, we adopt a standard data split: the training set includes 1610 videos, and the testing set contains 290 videos. As a larger-scale counterpart, \textbf{XD-Violence} \cite{xd-violence}, consists of 4754 untrimmed videos sourced from movies and YouTube, covering 6 categories of violent events. Of these, 3954 videos are utilized for training, and 800 videos for testing.

\noindent\textbf{Evaluation Protocol.} We assess the algorithm performance across both coarse-grained and fine-grained metrics. Specifically, for coarse-grained WS-VAD, we employ frame-level Average Precision (AP) as the metric for XD-Violence, and frame-level AUC for UCF-Crime. For fine-grained WS-VAD, consistent with prior works \cite{vadclip,itc}, we compute the mean Average Precision (mAP) over IoU thresholds ranging from 0.1 to 0.5 with an increment of 0.1. Besides, the average of these mAP values (denoted as AVG) is reported to provide a more comprehensive assessment.

\noindent\textbf{Implementation Details.}
We employ the pre-trained text encoder of CLIP with ViT-B/16, while multiple vision encoders, including I3D \cite{I3D}, C3D \cite{c3d}, and the CLIP visual encoder with ViT-B/16, are explored for frame feature extraction. GPT4 is adopted to extract anomaly attribute information. During training, the AdamW optimizer is used with a batch size of 64, a learning rate of 2e-5, and a total of 10 epochs. The thresholds $\theta_v$ and $\theta_s$ are set to 0.1 and 0.2. The value of $K$ is determined as $K = max(\lfloor T/16 \rfloor, 1)$, the momentum coefficient $\beta$ is 0.1, and $\tau$ is 0.9. 

\begin{table}[]
\renewcommand{\arraystretch}{1.05} 
\resizebox{0.48\textwidth}{!}{
\begin{tabular}{l|cccccc}
\toprule
\multirow{2}{*}{Methods}        & \multicolumn{6}{c}{mAP@IoU}   \\ 
\cline{2-7} 
& 0.1       & 0.2    & 0.3    & 0.4     & 0.5    & AVG    \\ 
\midrule
RAD    \cite{ucf}     & 5.73   & 4.41   & 2.69    & 1.93   & 1.44    & 3.24   \\
AVVD \cite{avvd}    & 10.27  & 7.01   & 6.25    & 3.42   & 3.29    & 6.05   \\
VadCLIP  \cite{vadclip}   & 11.72  & 7.83   & 6.40    & 4.53   & 2.93    & 6.68   \\
ITC   \cite{itc}      & 13.54  & 9.24   & 7.45    & 5.46   & 3.79    & 7.90    \\
ReFLIP \cite{reflip}  & 14.23  & 10.34  & 9.32    & 7.54   & 6.81    & 9.62   \\ 
LEC-VAD \cite{lecvad} &  19.65  &  17.17  & 14.37  &  9.45   &  7.18   & 13.56 \\
\rowcolor[HTML]{C0C0C0} 
LAS-VAD (Ours)  &  \textbf{22.07}  &  \textbf{19.96}   & \textbf{16.18}  &  \textbf{11.24}   &  \textbf{8.64}   & \textbf{15.62} \\
\bottomrule
\end{tabular}
} 
\vspace{-0.4em}
\caption{Fine-grained comparisons on UCF-Crime.}
\label{tab:fg_ucf}
\vspace{-1em}
\end{table}

\begin{table}[]
\centering
\renewcommand{\arraystretch}{1} 
\resizebox{0.48\textwidth}{!}{
    \begin{tabular}{ccc|cccccc}
    \toprule
    \multirow{2}{*}{ATT} & \multirow{2}{*}{ACC} & \multirow{2}{*}{IAM} & \multicolumn{6}{c}{mAP@IoU}                   \\ 
    \cline{4-9} 
     &    &     & 0.1   & 0.2   & 0.3   & 0.4   & 0.5   & AVG   \\ 
    \midrule
    \usym{2718}  & \usym{2718} & \usym{2718} & 36.86 & 29.75 & 23.00 & 17.49 & 14.11 & 24.24 \\
    \usym{2714}  & \usym{2718} & \usym{2718} & 39.15 & 32.36 & 25.47 & 19.02 & 16.50 & 26.50 \\
    \usym{2718} & \usym{2714} & \usym{2718} & 42.53 & 35.15 & 28.48 & 23.91 & 18.83 & 29.78 \\
    \usym{2718} & \usym{2718} & \usym{2714} & 42.91 & 35.64 & 28.46 & 23.86 & 19.05 & 29.98 \\
    \usym{2714} & \usym{2714} & \usym{2718} & 46.77 & 39.53 & 32.15 & 26.39 & 21.42 & 33.25 \\
    \usym{2714} & \usym{2718}  & \usym{2714} & 46.65 & 39.28 & 32.06 & 26.51 & 21.39 & 33.18 \\
    \usym{2718} & \usym{2714} & \usym{2714} & 48.13 & 41.96 & 33.70 & 27.95 & 22.69 & 34.89 \\
    \rowcolor[HTML]{C0C0C0} 
    \usym{2714}   & \usym{2714} & \usym{2714} & \textbf{50.18}  & \textbf{44.09}   & \textbf{36.00}   & \textbf{29.51}   & \textbf{24.65}  & \textbf{36.89} \\
    \bottomrule
    \end{tabular}
}
\caption{Ablation Studies on XD-Violence. ``IAM" denotes the intention awareness module, ``ACC" denotes the anomaly-connected components, and ``AAT" denotes anomaly attribute clues.}
\label{tab:ablation}
\vspace{-1.5em}
\end{table}

\subsection{Main Results}
We conduct a thorough evaluation of anomaly detection performance via benchmarking against prevalent methods on two datasets. Experimental results demonstrate that our LAS-VAD achieves state-of-the-art performance across all evaluation metrics and granularity levels. 

First, we evaluate coarse-grained anomaly detection across the XD-Violence and UCF-Crime datasets, with corresponding results reported in Tables \ref{tab:cg_xd} and Table \ref{tab:cg_ucf}. For XD-Violence, I3D and CLIP features are adopted, and both of them achieve the best performance. In detail, our LAS-VAD achieves 89.96 AP with I3D features and 87.92 AP with CLIP features, yielding respective absolute gains of 1.49 and 1.36 over LEC-VAD. Notably, our method even outperforms PE-MIL, which uses additional audio features alongside I3D features. For UCF-Crime, we employ C3D, I3D, and CLIP features, all of which achieve remarkable performance. In detail, our LAS-VAD achieves 86.04 AP with C3D features, yielding an absolute gain of 1.29 over LEC-VAD. Similarly, LAS-VAD attains 91.05 AP and 90.96 AP with I3D and CLIP features respectively, outperforming the SOTA $\pi$-VAD by 0.72 and LEC-VAD by 0.89 in absolute gain. These experimental results reveal that our model’s superiority when using diverse features.

Moreover, our LAS-VAD delivers strong performance in fine-grained WS-VAD. As displayed in Table \ref{tab:fg_xd}, it achieves substantial improvements across all metrics on XD-Violence, yielding an amazing AVG of 36.89, a 5\% improvement over LEC-VAD. For the more challenging UCF-Crime dataset, corresponding results are presented in Table \ref{tab:fg_ucf}. Our LAS-VAD maintains consistent and remarkable gains across all evaluation metrics, delivering a notable AVG of 15.62 and outperforming LEC-VAD by 15.2\%. These results further highlight our algorithm’s strength in distinguishing subtle differences among diverse anomalies.

\begin{figure}
\centering 
\includegraphics[width=0.46\textwidth]{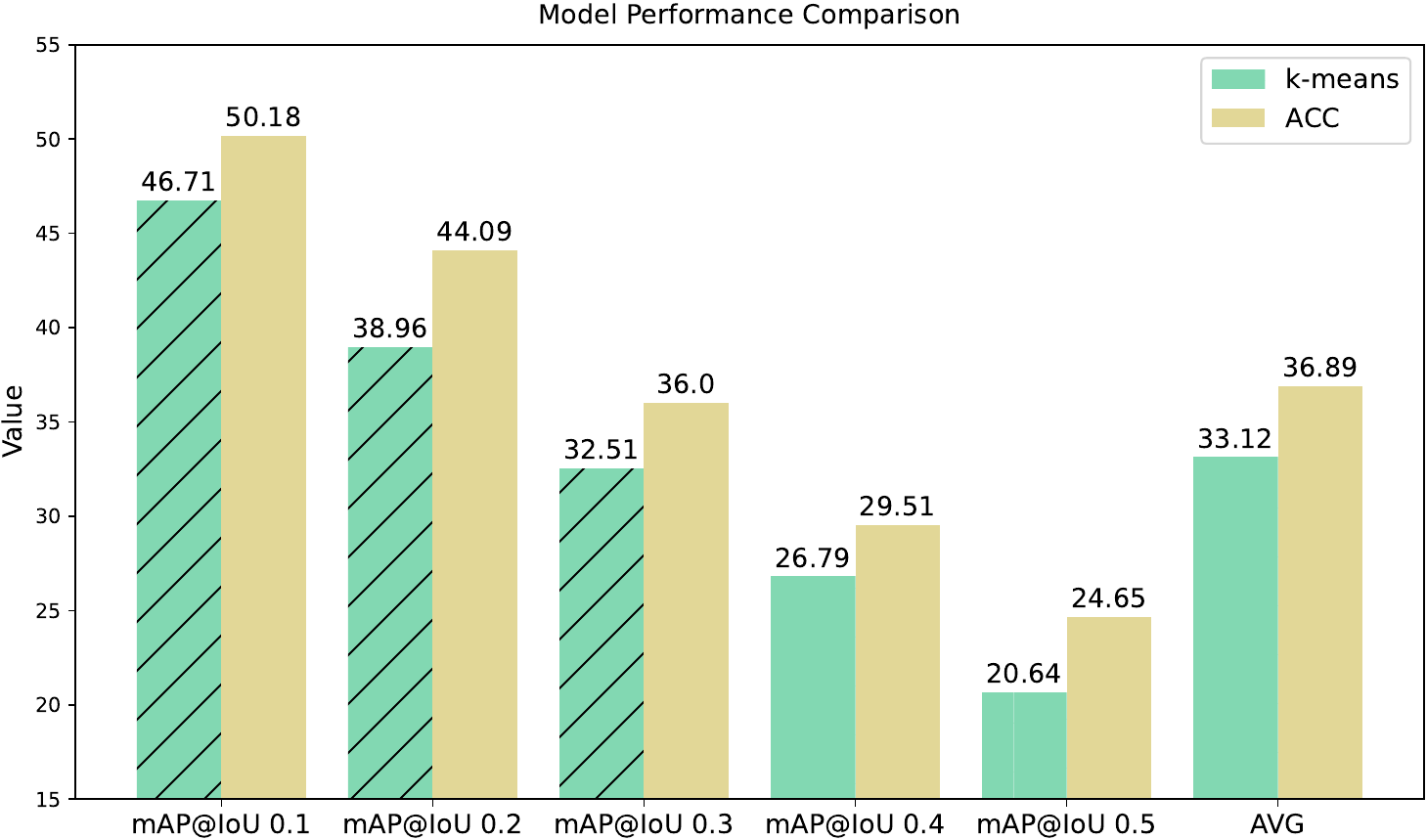} 
\caption{Ablation study on the utility of ACC.}
\label{fig:aba_acc} 
\vspace{-1.3em}
\end{figure}

\subsection{Ablation Study}
In this section, we perform comprehensive ablation studies on XD-Violence, systematically presenting more challenging fine-grained detection results to elucidate the contributions of individual factors to overall performance. 

\noindent\textbf{Analysis on model components.} We analyze the impact of our model’s components on detection performance, with results presented in Table \ref{tab:ablation}. In detail, we investigate the utilities of the proposed intention awareness mechanism (IAM), anomaly-connected components (ACC), and anomaly attribute clues (AAT) to clarify their individual contributions to the model’s overall performance. From Table \ref{tab:ablation}, it can be observed that integrating each component brings performance gains across all metrics, and the full model achieves the optimal performance over its castrated counterparts.

\begin{figure}
\centering
\begin{subfigure}[t]{0.22\textwidth}
    \centering
    \includegraphics[scale=0.2]{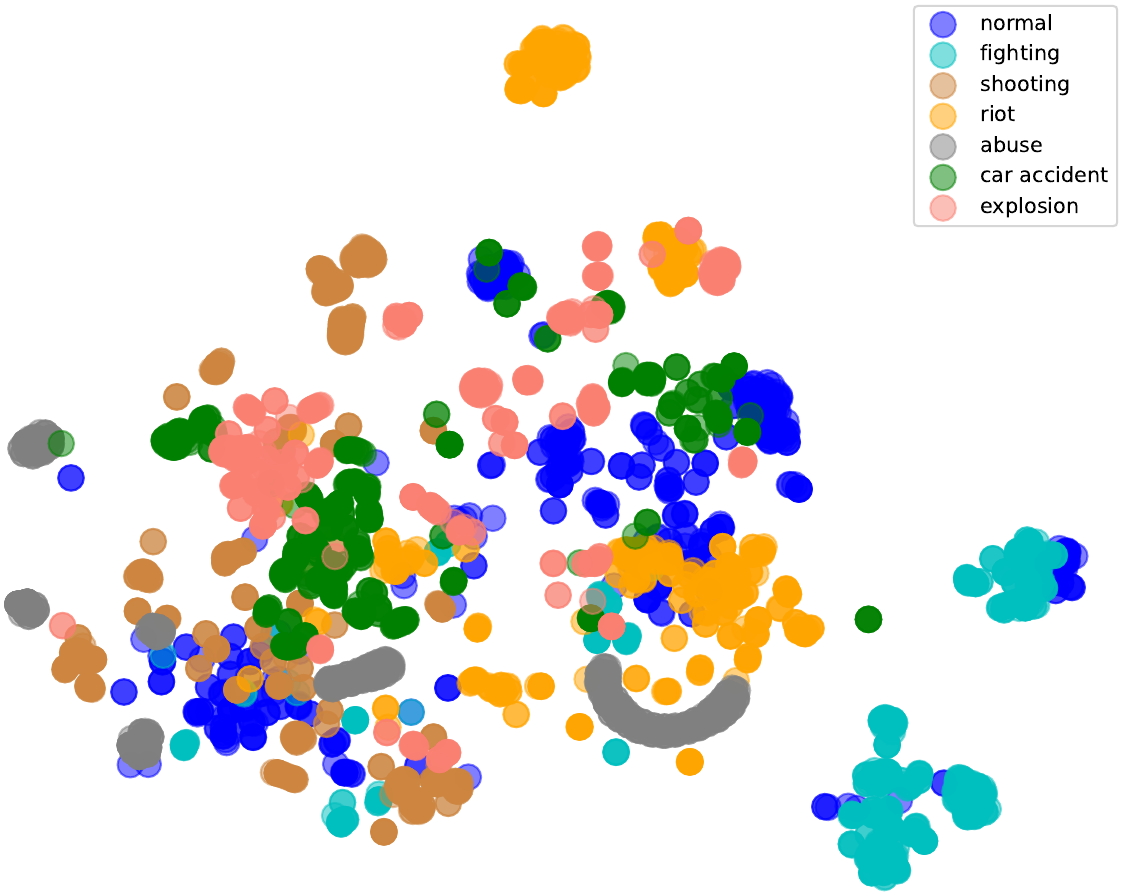}  
    \caption{CLIP}  
\end{subfigure}
\begin{subfigure}[t]{0.22\textwidth}
    \centering
    \includegraphics[scale=0.2]{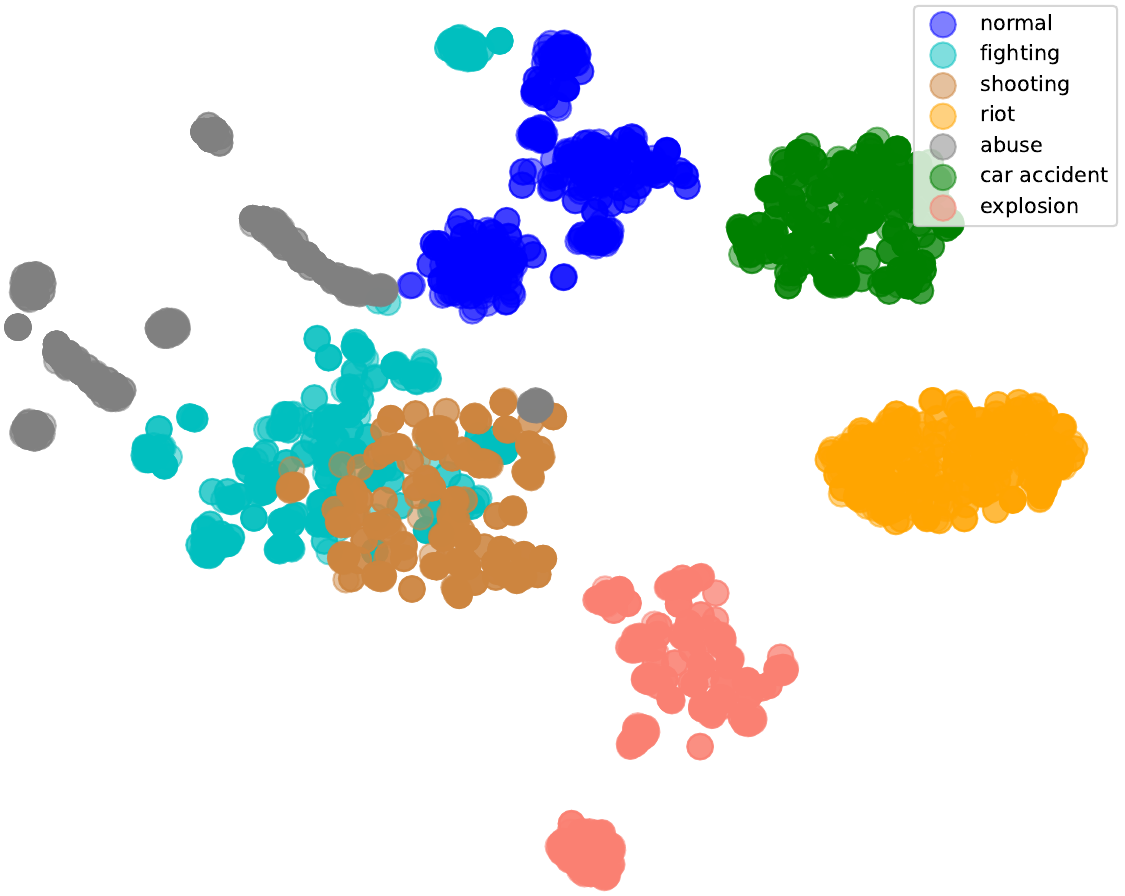}  
    \caption{k-means}  
\end{subfigure}

\begin{subfigure}[t]{0.22\textwidth}
    \centering
    \includegraphics[scale=0.2]{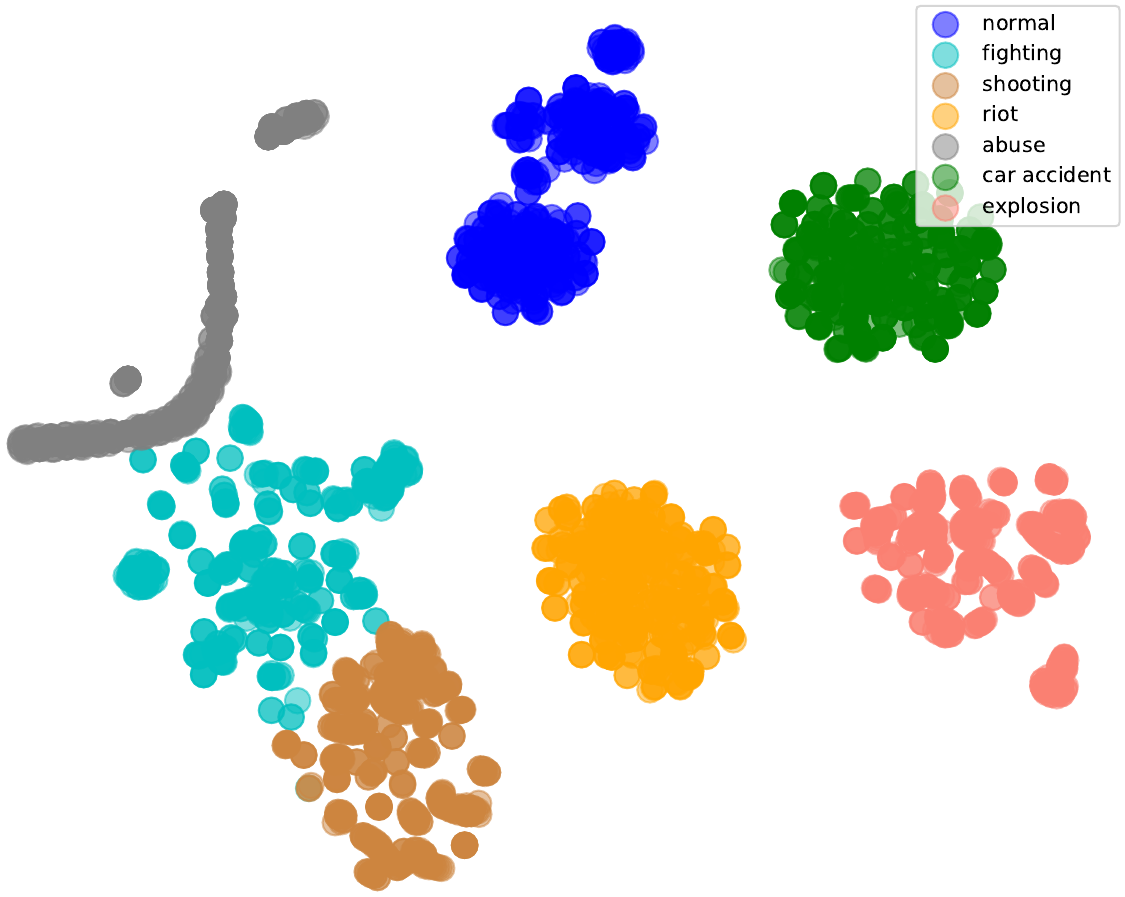}  
    \caption{ACC}  
\end{subfigure}
\begin{subfigure}[t]{0.22\textwidth}
    \centering
    \includegraphics[scale=0.19]{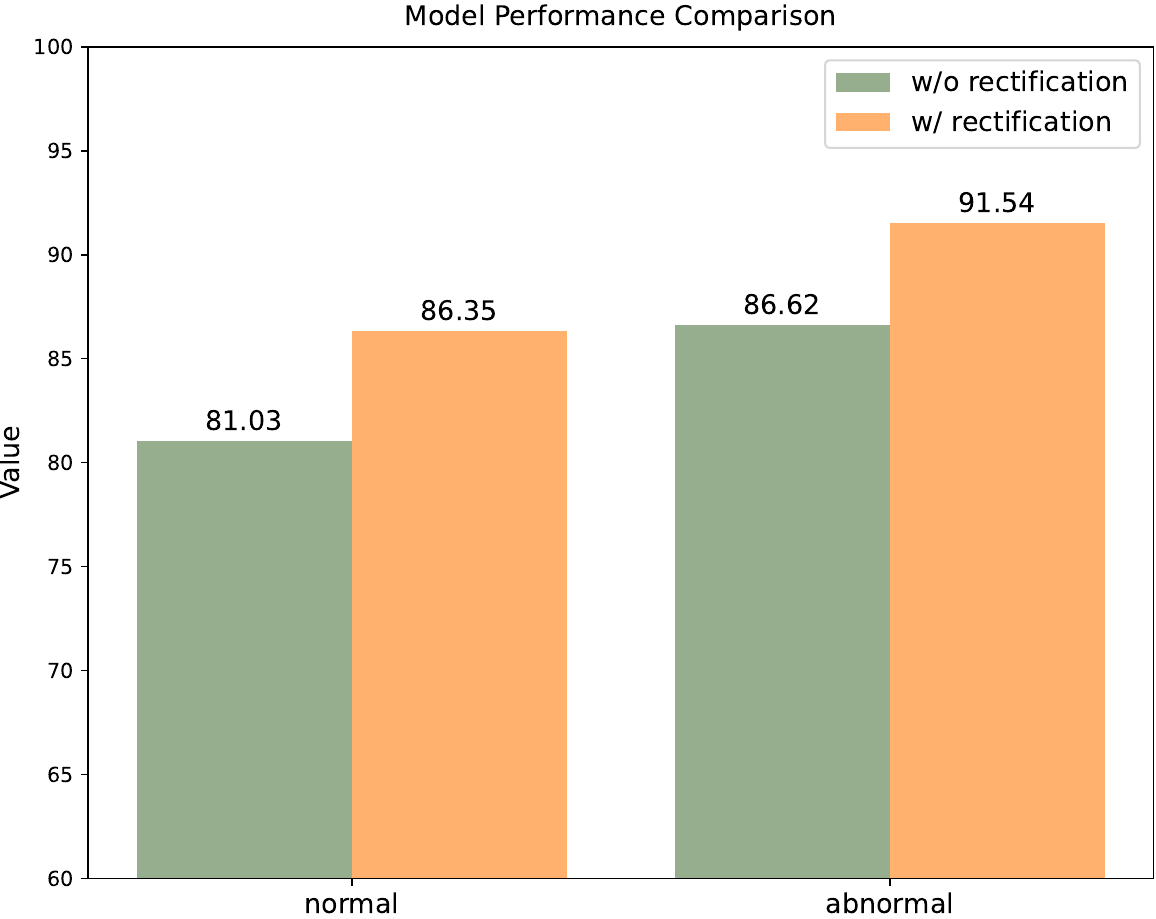}  
    \caption{Rectification mechanism}  
\end{subfigure}
\vspace{-0.5em}
\caption{Visualization of representations for different categories in (a), (b), and (c). (d) shows the utility of rectification mechanism.}
\label{fig:tsne}
\vspace{-1em}
\end{figure}

\noindent\textbf{Analysis on ACC.} To further validate ACC's utility, we adopt the k-means algorithm as an alternative to partition video frames into distinct groups. Notably, for the choice of $k$ in k-means, we distill the knowledge from ACC and set the number of components $r$ to $k$. Results are presented in Fig. \ref{fig:aba_acc}. Despite access to prior information of $k$ from ACC, k-means is observed to perform significantly worse than our ACC. We attribute this to the fact that ACC can model structural information among video frames via their adjacency matrix, whereas k-means only relies on feature similarity for partitioning. Besides, we investigate their capability in learning discriminative visual features, visualizing feature distributions from the original CLIP, the version with k-means, and our ACC-equipped model, and the results are presented in Fig. \ref{fig:tsne}. Features from our model form tighter, more sharply bounded clusters per category, indicating its capacity to learn discriminative features even without frame-level supervision.

Besides, we analyze the utility of rectification on the visual adjacent matrix $\mathcal{A}$. To validate this, we calculate recognition accuracy for normal and abnormal frames with and without the correction mechanism separately on test set. As shown in Fig. \ref{fig:tsne} (d), the correction mechanism improves performance for both normal and abnormal frames.

\begin{figure}
\centering 
\includegraphics[width=0.47\textwidth]{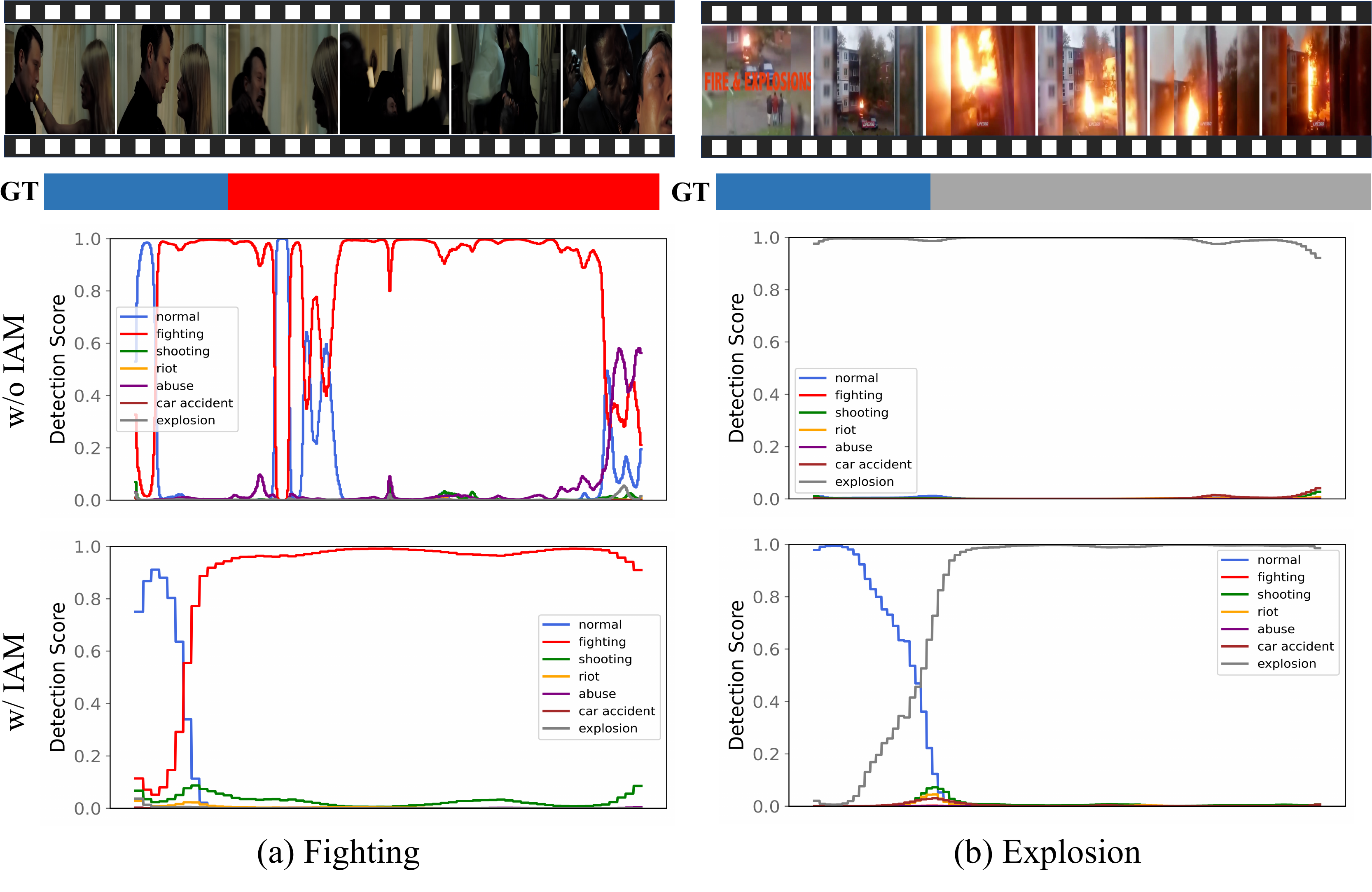} 
\vspace{-0.6em}
\caption{Fine-grained predictions w/ and w/o the proposed IAM.}
\label{fig:vis_ita} 
\vspace{-1em}
\end{figure}

\begin{table}[]
\renewcommand{\arraystretch}{1.05} 
\resizebox{0.48\textwidth}{!}{
\begin{tabular}{l|cccccc}
\toprule
\multirow{2}{*}{LLM}        & \multicolumn{6}{c}{mAP@IoU}   \\ 
\cline{2-7} 
& 0.1       & 0.2      & 0.3     & 0.4     & 0.5     & AVG    \\ 
\midrule
Lamma3       & 50.03  & 44.04  & 35.87   & \textbf{29.53}   & 24.58  & 36.81 \\
Qwen2         & 50.09  & 44.04  & 35.91   & 29.46   & 24.57  & 36.81 \\
Gemini2.5     & 50.14  & \textbf{44.11}  & 35.94   & 29.49   & 24.59  & 36.85 \\
DeepSeek V3  & 50.15  & 44.08  & 35.97   & 29.47   & 24.61  & 36.86 \\
GPT4      & \textbf{50.18}  & 44.09  & \textbf{36.00}   & 29.51   & \textbf{24.65}  & 36.89 \\
\bottomrule
\end{tabular}
} 
\vspace{-0.5em}
\caption{Different LLMs are tested on XD-Violence.}
\label{tab:aba_llm}
\vspace{-1.8em}
\end{table}

\begin{figure}
\centering
\begin{subfigure}[t]{0.23\textwidth}
    \centering
    \includegraphics[scale=0.27]{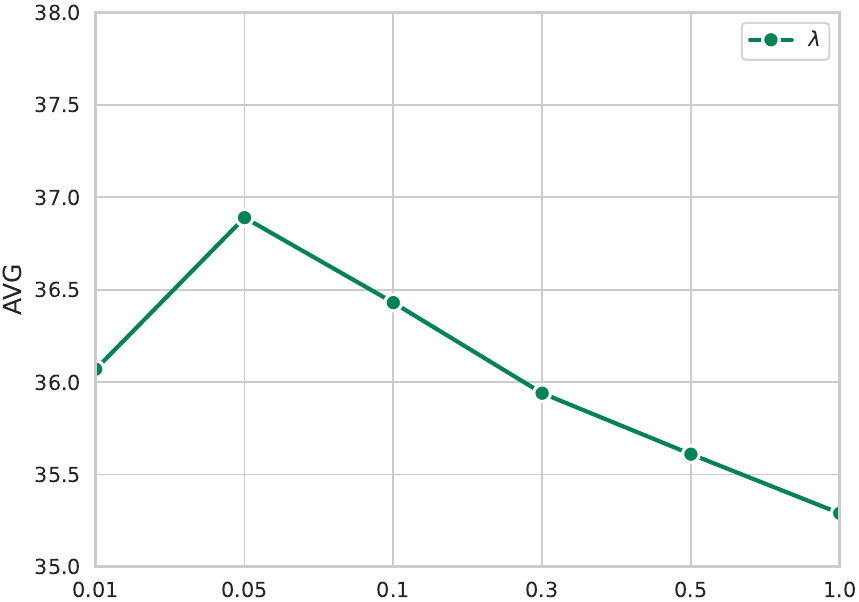}  
    \caption{$\lambda$}  
\end{subfigure}
\begin{subfigure}[t]{0.23\textwidth}
    \centering
    \includegraphics[scale=0.27]{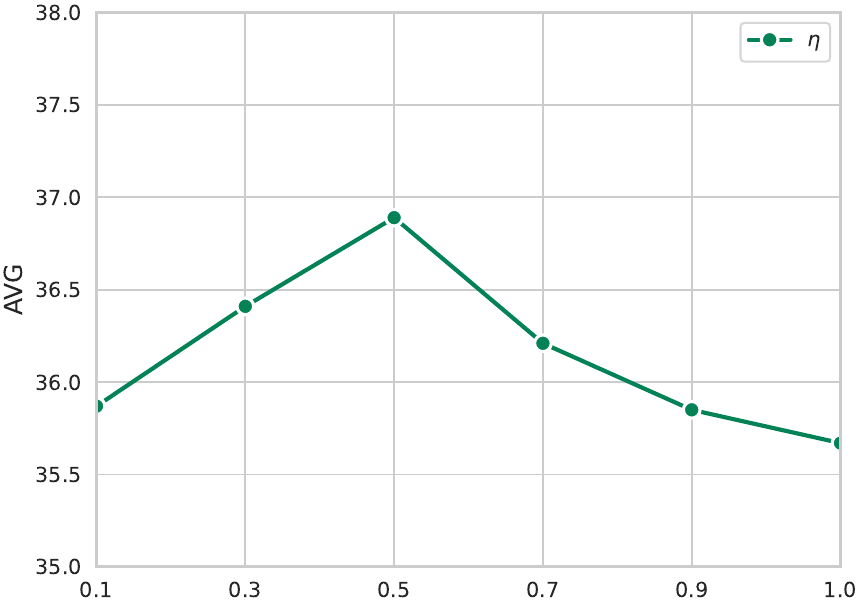}  
    \caption{$\eta$}  
\end{subfigure}
\vspace{-0.5em}
\caption{Effects of hyper-parameters.}
\label{fig:hp}
\vspace{-1.7em}
\end{figure}

\noindent\textbf{Analysis on IAM.} To investigate LAS-VAD's ablity in recognizing anomalous intentions, we compare some examples that adopt the version with IAM and without IAM, and the results presented in Fig. \ref{fig:vis_ita}. Compared with the model without the IAM module, the model integrated with the IAM module clearly perceives target intentions and critical events, such as rapid illegal physical contact (as shown in (a)) and rapid fire spread (as illustrated in (b)), thereby enabling more reasonable predictions.

\noindent\textbf{Analysis on anomaly attribute generation.} To generate attribute information of anomaly categories, we prompt LLMs to produce corresponding attribute descriptions. We compare performance across different LLMs, with results shown in Table \ref{tab:aba_llm}. Notably, different LLMs yield similar results, indicating that our model is insensitive to LLM types. GPT-4 is used as the default for attribute clue generation.

\noindent\textbf{Analysis on hyper-parameters.} We conduct an extensive exploration of the hyper-parameter configuration for $\lambda$ and $\eta$, with results presented in Fig. \ref{fig:hp}. For $\lambda$, we explore different values and get the optimal value at $\lambda=0.3$. For $\eta$, we incrementally adjust it from 0.1 to 1, and get the optimal value at $\eta=0.5$, which indicates that appropriately rectifying the visual adjacency matrix $\mathcal{A}$ via cross-modal information brings performance gains, while excessive rectification is counterproductive.


\section{Conclusion}
This paper proposed a novel LAS-VAD for WS-VAD to effectively learn anomaly semantics. LAS-VAD integrated an anomaly-connected component mechanism and an intention awareness mechanism. The former assigned video frames into distinct semantic groups where intra-group frames share consistent semantics, while the latter used an intention-aware strategy to distinguish similar normal and abnormal behaviors. Besides, we incorporated anomaly attribute clues to guide more accurate detection. 

\textbf{Acknowledge.} This work was supported in part by the National Natural Science Foundation of China under Grant 62406226, in part sponsored by Shanghai Sailing Program under Grant 24YF2748700, in part by New-Generation Information Technology under the Shanghai Key Technology R\&D Program under Grant 25511103500.

{
    \small
    \bibliographystyle{ieeenat_fullname}
    \bibliography{main}
}


\end{document}